\documentclass[a4paper]{article}
\usepackage[utf8]{inputenc}

\usepackage{INTERSPEECH2018}
\usepackage{multirow}

\title{Punctuation Prediction Model for Conversational Speech}

\name{Piotr \.Zelasko$^1$$^2$, Piotr Szyma\'nski$^1$$^3$, Jan Mizgajski$^1$, Adrian Szymczak$^1$, Yishay Carmiel$^1$, Najim Dehak$^4$}
\address{
  $^1$ Intelligent Wire, USA \\
  $^2$ Department of Computer Science, Electronics and Telecommunications, AGH University of Science and Technology, al. Mickiewicza 30, Krak\'ow, Poland \\
  $^3$ Department of Computational Intelligence, Wrocław University of Technology, Wybrzeże Stanisława Wyspiańskiego 27, 50-370 Wrocław, Poland \\
  $^4$ Center for Language and  Speech Processing, The Johns Hopkins University, Baltimore, MD, USA
}
\email{pzelasko@intelligentwire.com, piotr.szymanski@spoken.com, jan.mizgajski@spoken.com, adrian.szymczak@spoken.com, ycarmiel@intelligentwire.com, ndehak3@jhu.edu}

\begin{document}

\maketitle

\begin{abstract}
    \textbf{An ASR system usually does not predict any punctuation or capitalization.
    Lack of punctuation causes problems in result presentation and confuses both the human reader and off-the-shelf natural language processing algorithms.
    To overcome these limitations, we train two variants of Deep Neural Network (DNN) sequence labelling models - a Bidirectional Long Short-Term Memory (BLSTM) and a Convolutional Neural Network (CNN), to predict the punctuation.
    The models are trained on the Fisher corpus which includes punctuation annotation.
    In our experiments, we combine time-aligned and punctuated Fisher corpus transcripts using a sequence alignment algorithm.
    The neural networks are trained on Common Web Crawl GloVe embedding of the words in Fisher transcripts aligned with conversation side indicators and word time infomation.
    The CNNs yield a better precision and BLSTMs tend to have better recall.
    While BLSTMs make fewer mistakes overall, the punctuation predicted by the CNN is more accurate - especially in the case of question marks.
    Our results constitute significant evidence that the distribution of words in time, as well as pre-trained embeddings, can be useful in the punctuation prediction task.
}
\end{abstract}

\section{Introduction}\label{sec:introduction}

Automatic Speech Recognition (ASR) systems are becoming widely adopted in various applications, such as voice commands, voice assistants, dictation tools or conversation transcribers.
In many ASRs, a serious limitation is the lack of any punctuation or capitalization (with exception of some recent end-to-end models).
This can be problematic both in the case of visual presentation of the outputs, where the non-punctuated transcripts are confusing and difficult to read, and when these transcripts are used as inputs for downstream tasks such as those in the domain of Natural Language Processing (NLP).
Off-the-shelf NLP systems are usually trained on punctuated text, thus lack punctuation can cause a significant deterioration of their performance.

We are especially interested in addressing this issue in the domain of telephone conversational speech.
Our application transcribes telephone calls between customers and agents, and performs their semantic annotations to find particular and specific events, as well as an intents and moods of the interlocutors.
Providing punctuation became crucial for us to provide a high quality service.

Unlike many other machine learning tasks, punctuation prediction does not abound reference datasets that would enable supervised learning.
In principle any punctuated text source such as blogs, news articles or Wikipedia, could be used for training a punctuation prediction model, but most of them are hardly representative of the conversational language.
On the other hand, speech transcripts with proper punctuation are rather difficult to find or time-consuming to annotate.
In this work, we show that the English Fisher corpus~\cite{cieri2004fisher}, which contains about 11000 distinct conversations, can be successfully used to provide data for punctuation prediction.

To leverage the fact that we are working with conversational speech, we propose to use the recognition from both sides of the conversation to predict punctuation, as well as relative timing and duration of each word, which, to the best of our knowledge, has not been used before for punctuation prediction task.
Two variants of Deep Neural Network (DNN) sequence labelling models - a Bidirectional Long Short-Term Memory (BLSTM)  and a Convolutional Neural Network (CNN) were trained to predict the punctuation outputs for each word in the dialogue sequence.
Pre-trained GloVe~\cite{pennington2014glove} word embeddings were used with the intent of making the model more robust to different conversation topics than those that can be found in English Fisher corpus~\cite{cieri2004fisher}.
Both models achieve results that are on par with other work performed in this task for comparable domains.

The related research is presented in section~\ref{sec:relatedWork}.
Section~\ref{sec:methods} describes our approach to data preparation as well as model architectures and the details of their training.
We present and discuss the results in section~\ref{sec:results}.
Finally, we conclude our work in section~\ref{sec:conclusions}.

\section{Related work}\label{sec:relatedWork}

Early attempts focused on finding sentence boundaries ("dot prediction"), and for that purpose, several linguistic features were used: an n-gram language model, turn markers and parts of speech (POS) information~\cite{stolcke1996automatic}.
Subsequent research employed a maximum entropy model, which predicted dots, commas and question marks based on lexical features (words, n-grams and previous predictions) and prosodic features, represented as pause tokens of a specific length~\cite{huang2002maximum}.
It has been shown that the presence of pauses in speech can serve as an indicator of punctuation marks, but there is a significant variation in how different speakers use pauses~\cite{igras2016structure}.
Conditional Random Fields (CRF) based models were also proposed for this task~\cite{lu2010better,ueffing2013improved}.

Recently, an LSTM model with several variants has been proposed for this task, which similarly uses words and pauses tokens as inputs~\cite{tilk2015lstm,tilk2016bidirectional}.
The authors decided not to use additional prosodic features such as F0 or phone durations due to their subpar performance in~\cite{christensen2001punctuation}.
We wish to emphasize that relative word timing and duration have not been investigated by any of these works, and in principle, their fidelity should be higher than artificial, discretized pause tokens.

\section{Methods}\label{sec:methods}

\subsection{Data preparation}\label{subsec:dataPreparation}

Unlike other telephone speech corpora the Fisher corpus~\cite{cieri2004fisher} has properly punctuated transcripts.
While the most widely used version of the Fisher transcripts (available in LDC catalogue numbers LDC2004T19 and LDC2005T19) are the \textit{.txt} files containing time alignment, the majority of conversations also has a second transcript version in a \textit{.txo} file, which does not have time alignment, but has rich punctuation and proper capitalization.
The availability of this data provides an opportunity to utilize the information from both sides of the conversation to predict punctuation.

We represent a dialogue $\mathcal{W}$ as an ordered set $\mathcal{W} = \{w_i\}$ of words $w$, where each $w$ has several properties:
\begin{itemize}
    \item $t_i$ is the textual representation of word $w_i$;
    \item $c_i$ is a binary feature, representing which conversation side uttered word $w_i$;
    \item $s_i$ is a real number, describing time offset (in seconds) at which the word $w_i$ started;
    \item $d_i$ is a real number, describing the duration (in seconds) of the word $w_i$;
    \item $p_i$ is the punctuation symbol, which appears after word $w_i$.
\end{itemize}
The set is ordered on the $s$ property of each word, i.e. the starting time.
This formulation allows to elegantly represent interjections, interruptions and simultaneous speech, which are often encountered in dialogues.
The $p$ properties are only known at the training time and are being predicted during inference.
With this representation in mind, we treat the punctuation prediction problem as a sequence labelling task.

To fit the Fisher data into our model definition, we need to combine information from time-annotated and punctuated transcripts.
The first step is computing the forced alignment of the time-annotated transcripts to obtain word-level information about starting times and durations ($s$ and $d$ properties).
For that purpose, we used the Kaldi ASR toolkit~\cite{povey2011kaldi} with a LSTM-TDNN acoustic model trained with lattice-free Maximum Mutual Information (MMI) criterion~\cite{povey2016purely}.
In order to minimize the differences between two transcript versions we edited the Fisher data preparation script not to exclude single-word utterances and the text in parentheses, .

The next step is extraction of punctuation properties $p$ and conversation side properties $c$ from the punctuated transcripts.
We retain blanks (no punctuation), dots, commas and question marks.
Other punctuation classes were rejected (converted to blanks) due to their low frequency (e.g.\ exclamation marks or triple dots) or the fact that it is modeled by other properties of the representation (double dash - that marks an interruptions).

Finally, we combine the information obtained from both sources.
This task is not trivial, since both transcript versions may have slight differences.
We observed that this problem could be viewed as global alignment between two symbol sequences, which can be obtained by the application of the Needleman-Wunsch algorithm~\cite{needleman1970general}. The algorithm, originating in bioinformatics for DNA sequence alignment, is based on dynamic programming and is available in open-source Biopython library~\cite{bioinformaticsbtp163}.
We compute the alignment between two transcript versions separately for each channel in each recording and remove the words which appeared in only one of the transcripts.
Then, we concatenate the words from both channels into one sequence and sort it by the starting time $s$, which yields our dialogue representation.

\begin{figure}[t]
    \centering
    \includegraphics[width=0.5\linewidth]{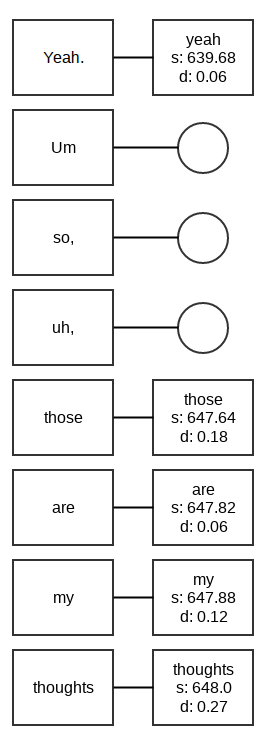}
    \caption{
    An example of alignment between two word sequences in Fisher: the time-annotated and the punctuation annotated.
    The $s$ stands for start time and $d$ stands for duration, both in seconds.
    The circles represent a blank symbol, i.e.\ no match for a given word in the second sequence.
    }
    \label{fig:sequence_alignment}
\end{figure}

\begin{table}[th]
    \caption{The total count of labels for each of the punctuation classes available in our training data set.}
    \label{tab:punctuationClasses}
    \centering
    \begin{tabular}{ r r r }
        \toprule
        \textbf{Class} & \textbf{Count} & \textbf{Percentage} \\
        \midrule
        blank         & 1429905 & 79.1\% \\
        comma         & 208289  & 11.5\% \\
        dot           & 148624  & 8.2\%  \\
        question mark & 22182   & 1.2\%  \\
        \bottomrule
    \end{tabular}

\end{table}

Since this is a sequence labelling task, we're predicting punctuation class for each word.
This results in a heavy class imbalance, as shown in table~\ref{tab:punctuationClasses}.
We attempted to mitigate this issue by introducing sample weighting based on predicted class frequency, however, it resulted in the model being skewed towards high recall, but much lower precision for the under-represented classes, manifesting as frequent false positives.

\subsection{Punctuation model}\label{subsec:punctuationModel}

\subsubsection{Features}\label{subsubsec:features}

There are several input features which we explored for our experiments.
The features which we used in every experiment are word embeddings and a conversation side indicator.
The word embeddings are 300-dimensional pre-trained GloVe~\cite{pennington2014glove} embeddings\footnote{
    The \textit{glove.42B.300d.zip} embeddings, which are available at https://nlp.stanford.edu/projects/glove.
}, trained on Common Web Crawl data.
Those weights are fixed during training.
We selected the embeddings for 50000 most frequent words.
Then we expanded this representation by added zeroes values to embed all out of vocabulary words to save GPU memory.
Increasing the vocabulary size to 100000 words did not provide any significant performance gains.
Additionally, we trained our own GloVe embeddings on conversational-like data (around 525M words) gathered by the University of Washington\footnote{
    The \textit{525M\_fisher\_conv\_web-filt+periods.gz} data set, which is available at https://ssli.ee.washington.edu/data.
} to investigate if these embeddings trained on conversational data would perform better, however, in some experiments, they resulted in either the F1 score being 0.2-0.3\% lower or a lack of model convergence.
We suspect this might be caused by a much smaller data quantity compared to the official GloVe embeddings.

The conversation side feature is a one-dimensional binary feature.

We used the word time information described by the interval between the start of the current word and start of the previous word, and duration of the current word, as features to the model.
We provided the interval instead of absolute offset time to obtain a more normal-like distribution for this feature.
Both of these features are speaker-adapted, i.e.\ they are standardized with regard to other words uttered by the same speaker in the same dialogue.
This also means that the pauses are not modelled explicitly as word tokens - they must be inferred by the model based on the subsequent word timings.

In some experiments, we used part of speech (POS) tags predicted by SpaCy~\footnote{https://spacy.io/}, although we didn't notice any significant improvement.
We hypothesize that either the POS tags did not introduce any predictive information, or that the performance of the tagger was poor in the absence of punctuation (and thus sentence segmentation).

\subsubsection{Architecture}\label{subsubsec:architecture}

We evaluated the performance of two types of models - one based on Convolutional Neural Nets (CNN), and the other based on Bidirectional Long Short-Term Memory (BLSTM) networks.
The input layer is a concatenation of the features described in~\ref{subsubsec:features}.
Both models were implemented using Keras~\cite{chollet2015keras} with Tensorflow~\cite{abadi2016tensorflow} backend.

The BLSTM model consists of four BLSTM layers, with each direction having 128 weights.
This model has the advantage of seeing a large context of words during training, and possibly the whole conversation during inference.

The CNN model uses several layers of 1D convolutions, which can be interpreted as fully-connected layers processing the input in small windows.
We additionally use dilated convolutions to broaden the context seen by each consecutive CNN layer.
Each layer is followed by a SELU activation~\cite{klambauer2017self}, which yielded a small improvement over batch normalization~\cite{ioffe2015batch} with ReLU~\cite{nair2010rectified}.
The setup which worked best for us is six 1D CNN layers, each with the filter size of 128 and padding which doesn't modify the word sequence length (i.e.\ \textit{same}).
The context width is equal to 3 for first five layers and equal to 20 for the last layer.
The middle four layers have a dilation rate of 2.

The final layer in both CNN and BLSTM model is fully-connected and followed by a softmax activation - this layer is applied separately at each time step to retrieve punctuation prediction for a given word.

To regularize the model we apply several measures:
\begin{itemize}
    \item a dropout layer with probability 0.5 before the softmax layer;
    \item 0.001 weight decay for the softmax layer weights and also for the BLSTM recurrent layers;
    \item we add Gaussian noise with standard deviation 0.1 to the time feature and embedding inputs, before the last softmax activation, and before SELU activations in the CNN model;
    \item SELU activations in the CNN model, which constrain the weights to a zero mean and unit variance distribution (which was verified by inspecting in TensorBoard).
\end{itemize}

\subsection{Training}\label{subsec:training}

To train the models, we use a standard, categorical cross-entropy loss function and the Adam optimizer~\cite{kingma2014adam} with default settings proposed by the authors.
The number of epochs is determined by early stopping, with two epochs patience.
We divide the Fisher conversations into training, validation and test sets with proportions 8:1:1.
To best utilize the GPU, we use a batch size of 256 and each sample in the batch is created by traversing the conversation in windows of 200 words.

\section{Results}\label{sec:results}

We present the results achieved by the CNN and BLSTM models with and without time features in table~\ref{tab:allResults}.
Each model is evaluated with precision, recall and F1 scores for each punctuation class separately.
We see that CNN models yield slightly higher precision for the punctuation classes, and BLSTM tends to have the better recall (and the inverse is true for the blank symbol).
Although the BLSTM model makes fewer mistakes overall, the punctuation predicted by the CNN model is more accurate - especially in the case of question marks.
The word-level time features yield minor improvement in both models, which suggests that the prosodic information carried by the relative word timing and their duration is useful in the punctuation prediction task.

\begin{table}[th]
    \caption{
        The per-class precision, recall and F1-score (in \%) achieved by the CNN and BLSTM models with pre-trained GloVe embeddings.
        All models used 300-dimensional word embeddings and 1-dimensional boolean conversation side features, and the +T models additionaly used two 1-dimensional time features.
        The $\epsilon$ symbol denotes a blank prediction.
    }
    \label{tab:allResults}
    \centering
    \begin{tabular}{ r r r r r }
        \toprule
        \textbf{Model} & \textbf{Class} & \textbf{Precision} & \textbf{Recall} & \textbf{F1} \\
        \midrule
        \multirow{4}{*}{CNN}
        & $\epsilon$ & 91.7 & \textbf{95.5} & 93.5 \\
        & . & 67.7 & 58.6 & 62.8 \\
        & ? & 70.8 & 45.1 & 55.1 \\
        & , & 68.3 & 58.1 & 62.8 \\
        \midrule
        \multirow{4}{*}{CNN+T}
        & $\epsilon$ & 92.3 & 95.2 & 93.8 \\
        & . & \textbf{68.6} & 63.3 & 65.9 \\
        & ? & \textbf{72.9} & 46.7 & 57.0 \\
        & , & \textbf{68.7} & 60.3 & 64.2 \\
        \midrule
        \multirow{4}{*}{BLSTM}
        & $\epsilon$ & 92.7 & 94.9 & 93.8 \\
        & . & 66.9 & 63.1 & 64.9 \\
        & ? & 70.2 & 47.3 & 56.5 \\
        & , & 67.9 & 61.8 & 64.7 \\
        \midrule
        \multirow{4}{*}{BLSTM+T}
        & $\epsilon$ & \textbf{93.5} & 94.7 & \textbf{94.1} \\
        & . & 67.9 & \textbf{66.7} & \textbf{67.3} \\
        & ? & 64.7 & \textbf{54.6} & \textbf{59.2} \\
        & , & 68.2 & \textbf{64.1} & \textbf{66.1} \\
        \bottomrule
    \end{tabular}
\end{table}

\begin{figure}[t]
    \centering
    \includegraphics[width=\linewidth]{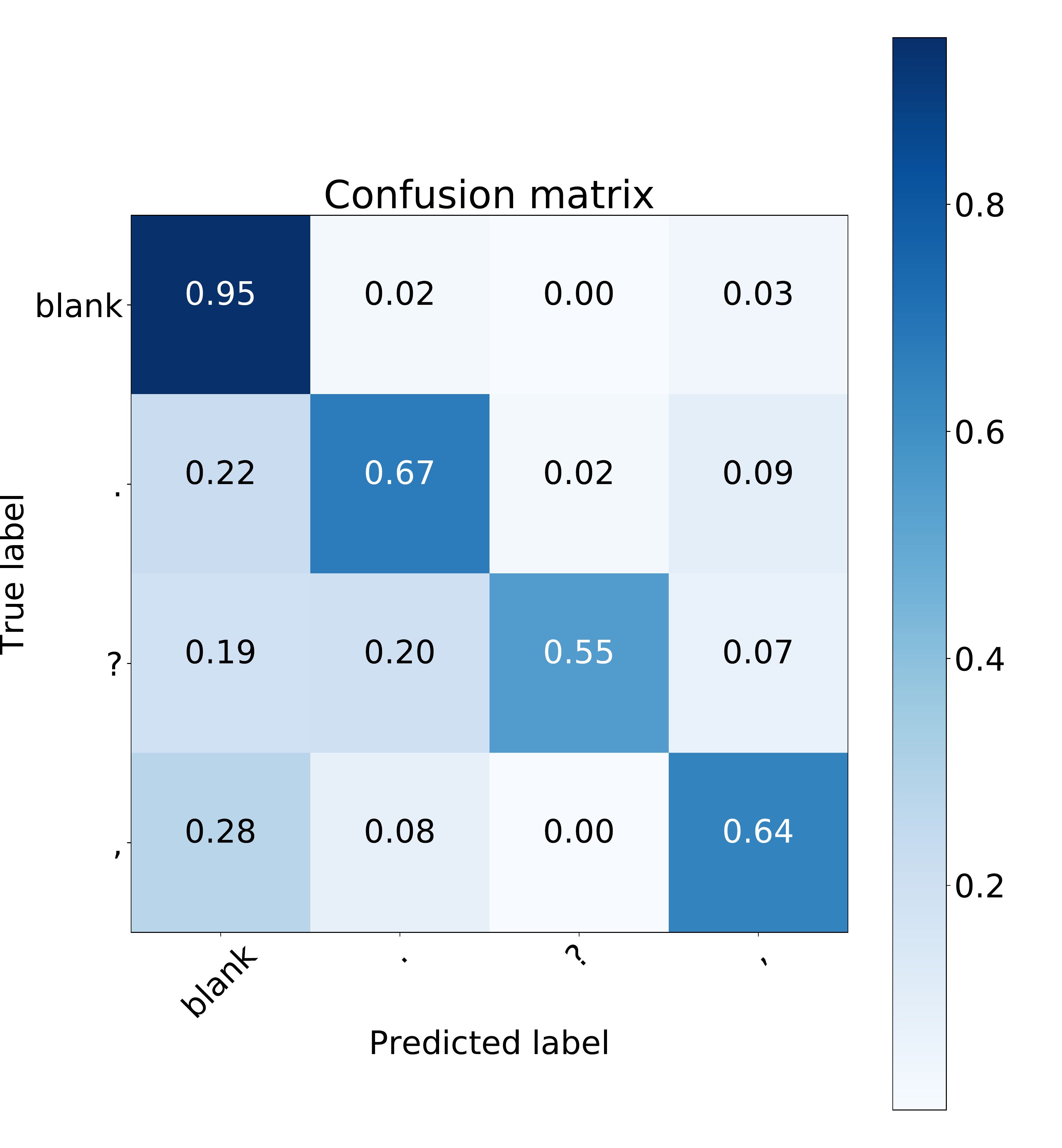}
    \caption{
        Confusion matrix for the BLSTM+T model, normalized with regard to true labels (i.e. rows).
    }
    \label{fig:confusionMatrix}
\end{figure}

For the BLSTM+T model we show the confusion matrix in figure~\ref{fig:confusionMatrix}.
This matrix is row-normalized to better illustrate per-class mistakes, but the reader should note that due to the class imbalance (shown in table~\ref{tab:punctuationClasses}), this confusion matrix is almost symmetric regarding absolute numbers.

We observe several interesting types of mistakes.
First of all, the blanks and commas are most frequently confounded types (around 55k false positives and 44k false negatives), which in our opinion is the least harmful type of mistake, given that the placement of commas in transcribed speech can often be arbitrary.
All of the punctuation classes labels are missed about 20\% of the time (i.e. blank is predicted) relatively to their occurrence count.
The question mark is the most difficult class to predict and is often mistaken with the dot (about 20\% of question marks), relatively rarely inserted in place of any other class.
This can most likely be explained by the scarcity of labels for this class.

Below is an example part of a Fisher dialogue showcasing the predictions of the punctuation model.
Note: words start with a capital letter only after a dot appears.

\begin{quote}
L: Oh, and that's west paterson. I don't know

R: Oh,

L: if

R: okay.

L: that counts.

R: Okay. Okay. Yeah, west peterson is nice.
    [laughter] So, i didn't even understand the ah,
    the topic of the day did you hear it?

L: I [noise] i heard first i heard censorship.
    And then i heard, ah, today's topic is something
    about public schools. It was i think, ah,
    should public schools

R: Do something about books

L: be allowed

R: kids

L: to censor

R: read?

L: certain books.
\end{quote}

Besides the quantitative evaluation, we also performed a qualitative investigation of the predictions of both models on the ASR transcripts of calls from a different domain than Fisher.
Since we do not have the golden labels for this data, this evaluation is highly subjective.
We observed that the CNN model tends to yield less confusing mistakes and outputs transcripts with higher, subjective readability, which is supported by the higher precision scores obtained by this model.
We suspect that this effect is amplified by the fact that the BLSTM model is more vulnerable to ASR mistakes due to the larger context size during inference.

\section{Conclusions}\label{sec:conclusions}

We presented two kinds of punctuation predictions DNN models - BLSTM and CNN based - which operate on a conversation, represented as a sequence of words, and utilize word embeddings, conversation side and per-word timing information as features.
We used two versions of the Fisher corpus transcripts - time-aligned and punctuated - along with sequence alignment procedure to procure the training and evaluation data.
Our results constitute significant evidence that the distribution of words in time, as well as pre-trained word embeddings, can be useful in the punctuation prediction task in the domain of conversational speech.
We've shown that the CNN architecture tends to achieve better precision scores, while the BLSTM variant is characterized by overall better recall and F1 measure.
These models can be easily applied in a production environment to provide punctuation annotations for speech recognition system transcripts, where all of the model input features are available.
For the future work, we'd like to investigate how much improvement can be gained by using prosodic features, as well as more sophisticated neural network architectures, such as models with attention~\cite{chan2016listen}.

\bibliographystyle{IEEEtran}

\bibliography{punctuation}

\end{document}